\documentclass[10pt,twocolumn,letterpaper]{article}

\usepackage{iccv}              

\definecolor{iccvblue}{rgb}{0.21,0.49,0.74}
\usepackage[pagebackref,breaklinks,colorlinks,allcolors=iccvblue]{hyperref}
\usepackage{multirow}
\usepackage[marginal]{footmisc}
\usepackage{hyperref}
\hypersetup{urlcolor = magenta}
\def\confName{ICCV}
\def\confYear{2025}

\title{Structure-Guided Diffusion Models for High-Fidelity Portrait Shadow Removal}

\author{
\begin{tabular}{@{}c@{}}
Wanchang Yu\textsuperscript{1} \hspace{0.3em} \hspace{0.3em} 
Qing Zhang\textsuperscript{1,2}\thanks{Corresponding author} \hspace{0.3em} \hspace{0.3em} 
Rongjia Zheng\textsuperscript{1} \hspace{0.3em} \hspace{0.3em} 
Wei-Shi Zheng\textsuperscript{1,2}
\end{tabular} \\
\textsuperscript{1}School of Computer Science and Engineering, Sun Yat-sen University, China \\
\textsuperscript{2}Key Laboratory of Machine Intelligence and Advanced Computing, Ministry of Education, China\\
}
\definecolor{ywc}{rgb}{0.0, 0.4, 0.2}

\begin{document}
\twocolumn[{
	\renewcommand\twocolumn[1][]{#1}
	\maketitle
    \vspace*{-8mm}
	\begin{center}
		\captionsetup{type=figure}
        \captionsetup[subfigure]{labelformat=empty}
		\begin{minipage}[b]{1.00\linewidth}
       \begin{subfigure}[c]{0.163\textwidth}
            \includegraphics[width=1.0\textwidth]{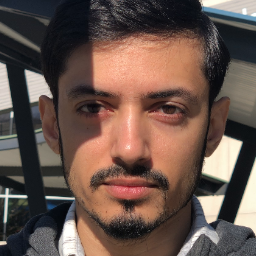}\vspace{1pt}
            \includegraphics[width=1.0\textwidth]{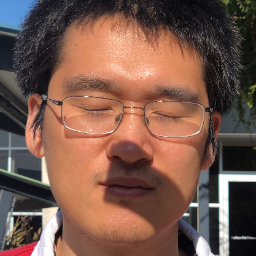}
            \caption{Input}
        \end{subfigure}
        \begin{subfigure}[c]{0.163\textwidth}
            \includegraphics[width=1.0\textwidth]{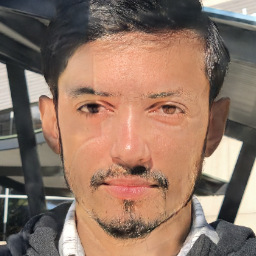}\vspace{1pt}
            \includegraphics[width=1.0\textwidth]{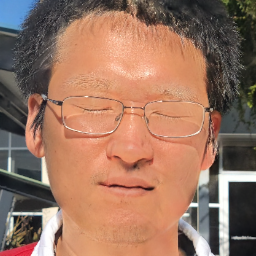}
            \caption{ShdowDiffusion~\cite{guo2023shadowdiffusion}}
        \end{subfigure}
        \begin{subfigure}[c]{0.163\textwidth}
            \includegraphics[width=1.0\textwidth]{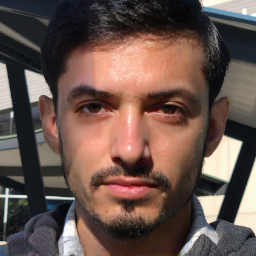}\vspace{1pt}
            \includegraphics[width=1.0\textwidth]{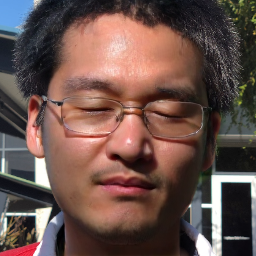}
            \caption{IC-light~\cite{zhangscaling}}
        \end{subfigure}
        \begin{subfigure}[c]{0.163\textwidth}
            \includegraphics[width=1.0\textwidth]{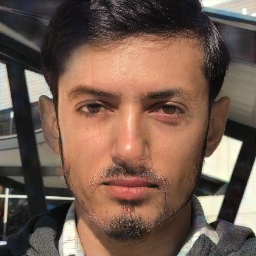}\vspace{1pt}
            \includegraphics[width=1.0\textwidth]{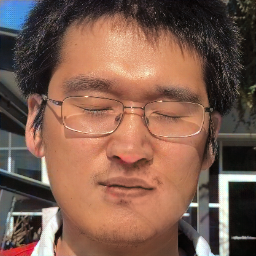}
            \caption{BSR~\cite{liu2022blind}}
        \end{subfigure}
        \begin{subfigure}[c]{0.163\textwidth}
            \includegraphics[width=1.0\textwidth]{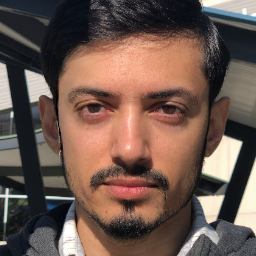}\vspace{1pt}
            \includegraphics[width=1.0\textwidth]{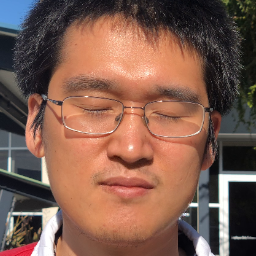}
            \caption{Ours}
        \end{subfigure}
        \begin{subfigure}[c]{0.163\textwidth}
            \includegraphics[width=1.0\textwidth]{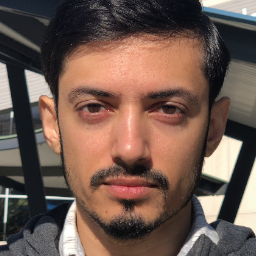}\vspace{1pt}
            \includegraphics[width=1.0\textwidth]{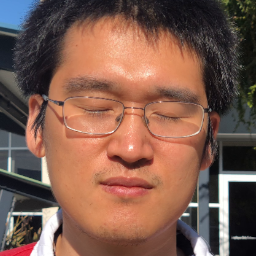}
            \caption{GT}
        \end{subfigure}
    \end{minipage}
    \caption{\textbf{Comparison on portrait shadow removal.} Our method allows to robustly remove facial cast shadows while faithfully preserving the portrait identities as well as structure details in shadow regions and also ensuring the overall naturalness.}
    \label{fig:teaser}
    \end{center}
    }]

\renewcommand{\thefootnote}{}
\footnotetext{$^*$Corresponding author (zhangq93@mail.sysu.edu.cn).}
\begin{abstract}
We present a diffusion-based portrait shadow removal approach that can robustly produce high-fidelity results. Unlike previous methods, we cast shadow removal as diffusion-based inpainting. To this end, we first train a shadow-independent structure extraction network on a real-world portrait dataset with various synthetic lighting conditions, which allows to generate a shadow-independent structure map including facial details while excluding the unwanted shadow boundaries. The structure map is then used as condition to train a structure-guided inpainting diffusion model for removing shadows in a generative manner. Finally, to restore the fine-scale details (e.g., eyelashes, moles and spots) that may not be captured by the structure map, we take the gradients inside the shadow regions as guidance and train a detail restoration diffusion model to refine the shadow removal result. Extensive experiments on the benchmark datasets show that our method clearly outperforms existing methods, and is effective to avoid previously common issues such as facial identity tampering, shadow residual, color distortion, structure blurring, and loss of details. Our code is available at \href{https://github.com/wanchang-yu/Structure-Guided-Diffusion-for-Portrait-Shadow-Removal}{https://github.com/wanchang-yu/Structure-Guided-Diffusion-for-Portrait-Shadow-Removal}.

\end{abstract}

\section{Introduction}
Taking selfies is very popular nowadays due to the readily-available cameras on various devices, particularly the mobile phones. However, portrait photos taken in the wild lacking professional lighting conditions often suffer from undesired foreign shadows cast by external objects (see Figure~\ref{fig:teaser} for examples). The presence of facial shadows not only severely degrade the overall visual appeal of the portraits, but also adversely affect many downstream vision tasks, such as face detection and recognition~\cite{dong2022deep, drozdowski2021watchlist,zhang2020refineface,wang2021representative}, and portrait image editing~\cite{jiang2021talk,xu2022transeditor}. Hence, portrait shadow removal techniques are widely required.

Currently, portrait shadow removal remains a challenge despite the remarkable progress in natural image shadow removal. The reason behind is twofold. First of all, due to factors such as privacy and difficulty in data collection, large-scale paired data for portrait shadow is extremely scarce, which significantly limits the development of portrait shadow removal methods, especially learning-based solutions. On the other hand, as this task is portrait-oriented, it has very low tolerance for visual artifacts, even minor imperfections on facial color and details could make the final result unnatural and unacceptable. Note, although existing software allow users to interactively adjust photos, it is still rather tedious and difficult for non-experts to manually produce high-fidelity portrait shadow removal results.

Existing shadow removal methods~\cite{guo2023shadowdiffusion,li2023leveraging,xiao2024homoformer} mainly target natural images due to the availability of large-scale paired datasets. However, these methods basically fail to deal with portrait shadows, as shown in Figures~\ref{fig:com_psm_data} and \ref{fig:com_other_data}. As there is no accessible large-scale paired datasets for portrait shadow removal, current methods in this field focus primarily on constructing synthetic paired data to enable supervised learning of shadow removal~\cite{zhangscaling,zhang2020portrait,liu2022blind,lyu2022portrait}. Suffering from the domain gap between real-world portraits and synthetic data, these methods typically do not work well for real-world portraits with complex shadows. To avoid the domain gap issue led by synthetic training data, a zero-shot solution is introduced in~\cite{he2021unsupervised}, which proposed to leverage the generative prior from pre-trained StyleGAN2~\cite{karras2020analyzing} to recover shadow-free portrait through test-time GAN inversion. However, this method tends to modify face identity and incur unnatural results. Note, although recent image relighting techniques \cite{weir2022deep,futschik2023controllable,hou2024compose,yoon2024generative} have shown remarkable progress, they struggle to enable high-fidelity portrait shadow removal due to the inherent difficulty in estimating ambient light and facial geometry from a single image.



To robustly produce high-fidelity portrait shadow removal results, we in this work present a diffusion-based approach. Particularly, we propose to cast portrait shadow removal as diffusion-based inpainting, so as to take full advantage of the strong generation capability of diffusion models. To this end, we first train a shadow-independent structure extraction network (SE-Net) on a dataset of paired real-world portraits with various synthetic lighting conditions, from which we obtain a structure map containing facial details while excluding the shadow boundaries. With the structure map, we then train a structure-guided inpainting diffusion model by enforcing reconstruction of randomly masked shadow-free portraits. Finally, to restore the fine-scale facial details that may not be captured by the structure map, we further refine the shadow removal result by a detail restoration diffusion model under the guidance of the gradients in the original shadow regions. Note, although our method requires a test shadow mask as input, it is highly robust to inaccurate masks, requiring only that the mask covers the shadow regions, even if it includes non-shadow regions. The major contributions of this work are:

\begin{itemize}
    \item We present a simple yet effective unsupervised approach that addresses portrait shadow removal from a perspective of diffusion-based inpainting.
    \item We introduce a relighting-based data synthesis strategy for constructing the training data, and develop a shadow-independent structure extraction network.
    \item We show that our method outperforms state-of-the-art methods for portrait shadow removal. 
    
\end{itemize}

\section{Related Work}
\textbf{Shadow removal.} Early methods typically rely on various handcrafted priors or assumptions to work~\cite{finlayson2009entropy,gryka2015learning,guo2012paired}. Recent methods are basically deep learning based \cite{hu2019mask,jin2021dc,liu2021shadow,liu2024recasting,guo2023shadowformer,guo2023shadowdiffusion,jin2024des3,zhu2022bijective,zhu2022efficient,cun2020towards,le2019shadow}, benefiting mostly from the availability of large-scale paired datasets such as ISTD~\cite{wang2018stacked} and SRD~\cite{qu2017deshadownet}. However, as these methods are tailored for natural images, they are not so effective in dealing with portraits. In contrast to natural images, portrait shadow removal has received much less research attention, due to the lack of large-scale paired datasets. To overcome the data scarcity issue, some methods~\cite{zhang2020portrait,liu2022blind} proposed to synthesize paired shadow and shadow-free portrait images for training. As their data synthesis strategies involve only simple adjustments on brightness and saturation, without considering facial geometry and albedo, their methods typically do not work well for real-world portraits. He et al.~\cite{he2021unsupervised} achieved zero-shot portrait shadow removal by utilizing generative prior from StyleGAN2, but this method may result in undesired modifications to facial details and thus fails to preserve the face identity. More recently, image relighting is also used for shadow removal~\cite{weir2022deep,zhangscaling,futschik2023controllable,hou2024compose,yoon2024generative}.However, this method tends to induce unnatural results because it is very challenging to obtain accurate prediction of diffuse lighting and albedo from a single portrait image.

\vspace{0.5em}
\noindent \textbf{Image inpainting.} Before the era of deep learning, patch-based self-similarity of natural images has long been the foundation for image inpainting~\cite{criminisi2004region,wexler2007space}. The emergence of deep learning has then led to great progress in image inpainting~\cite{pathak2016context,iizuka2017globally,yu2018generative,suvorov2022resolution}, as it is quite easy to construct paired training data. Recently, diffusion model has been a preferable choice for image inpainting~\cite{lugmayr2022repaint,saharia2022palette,corneanu2024latentpaint,xie2023smartbrush,zhu2024text,ji2024diffusion}, due to its impressive generation capability. Note, naive applying diffusion models to portrait shadow removal fail to produce visually natural results with unchanged portrait identity. A recent work closely related to our method also shows that it is feasible to remove shadows via image inpainting~\cite{li2023leveraging}. However, it does not work well for portrait shadow removal (see Figures~\ref{fig:com_psm_data} and \ref{fig:com_other_data}), even we fine-tuned their model on real-world paired shadow and shadow-free portrait images provided in \cite{zhang2020portrait}.

\begin{figure*}[t]
    \centering
    \includegraphics[width=\linewidth]{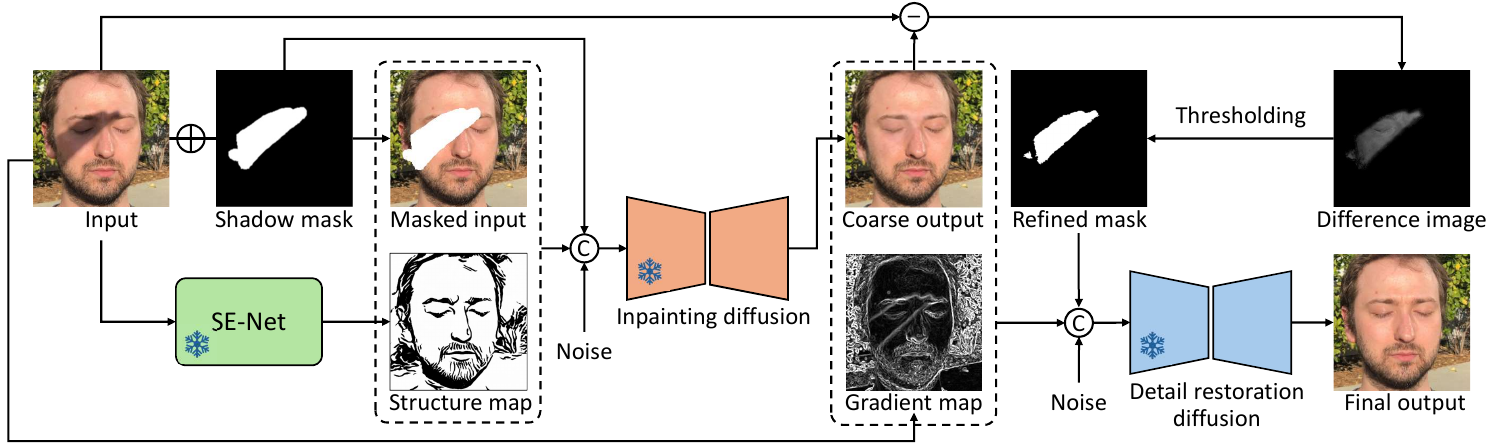} \\
    \vspace{-2mm}
\caption{\textbf{Inference pipeline of our method.} Given a input portrait image and the corresponding shadow mask, we first feed the input image to the shadow-independent structure extraction network (SE-Net) to get a structure map. Then, we perform shadow removal by inpainting the shadow regions with a diffusion model conditioned on the structure map. Finally, the shadow removal output is refined by a detail restoration diffusion model to restore the fine-scale facial details that may not captured by the strcutre map.}
\label{fig_inference}
\end{figure*}

\begin{figure}[t]
    \centering
    \includegraphics[width=\linewidth]{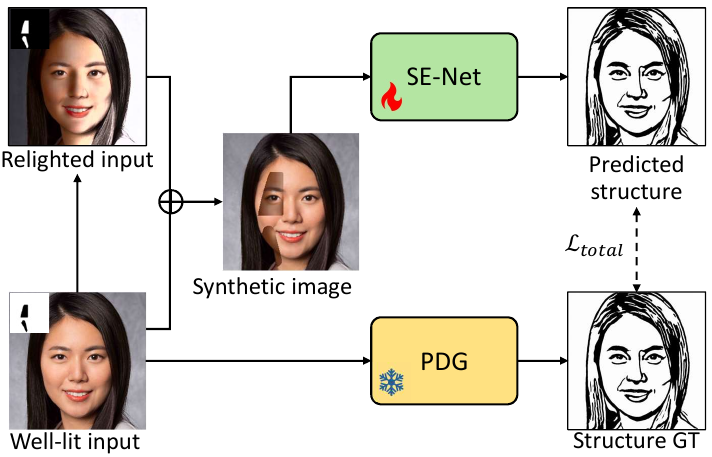}\\
    \vspace{-2mm}
\caption{\textbf{Training pipeline of SE-Net}. Note, we adopt a trained PDG model for structure extraction from \cite{yi2020unpaired}, while it is not invariant to illumination changes.}
\label{fig_SE}
\end{figure}
\section{Our Method}
Given a portrait image with cast shadows, our method aims to remove the facial shadows while faithfully preserving the facial details in the shadow regions. Figure~\ref{fig_inference} presents the overall inference pipeline of our method.


\begin{figure*}[t]
    \centering
    \includegraphics[width=\linewidth]{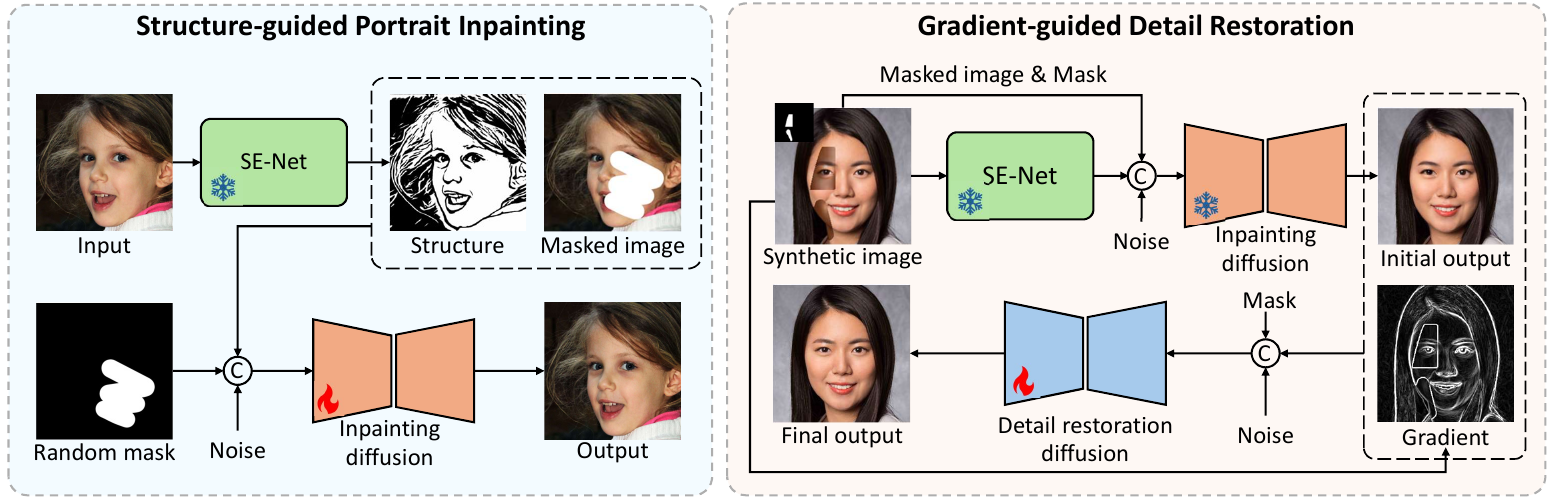} \\
    \vspace{-2mm}
\caption{\textbf{Training pipeline of our diffusion models.} Here we introduce how our structure-guided portrait inpainting diffusion model and gradient-guided detail restoration diffusion model are trained.}
\label{fig_train_diff}
\end{figure*}

\subsection{Shadow-Independent Structure Extraction} \label{SE}
{As shown in Figure~\ref{fig_inference}, the key to obtaining high-quality portrait shadow removal through inpainting is to acquire a structure map free of shadow boundaries for guiding the inpainting process. However, although existing structure/edge extraction methods are quite effective in identifying edges and structures, they are sensitive to illumination discontinuities and will inevitably include shadow boundaries in their results (see Figure~\ref{fig:different_structures}). Therefore, we propose to build a shadow-independent structure extraction network.

\vspace{0.5em}
\noindent\textbf{Training data synthesis.} To enable effective training of the shadow-independent structure extraction network, as shown in Figure~\ref{fig_SE}, we construct a synthetic paired dataset of well-lit real-world portraits without foreign shadows on faces and the corresponding relighted portraits with various illumination discontinuities, based on the shadow-free portraits in the CelebA dataset~\cite{liu2015deep}. Specifically, for a well-lit portrait $I$, we first apply an off-the-shelf physics-based relighting method \cite{zhou2019deep} over it to get a relighted image $I_{relit}$, which is then combined with the original portrait $I$ to produce a synthetic image $I_{syn}$ with illumination discontinuities by 
\begin{equation}
I_{syn}=M\odot I_{relit}+(1-M)\odot I,
\label{eq2}
\end{equation}
where $M$ is a random facial mask, and $\odot$ denotes element-wise multiplication. Note, our synthetic data is only used for training the SE-Net and the subsequent detail resotration diffusion model, rather than for directly training shadow removal model as done in \cite{zhang2020portrait}.

\vspace{0.5em}
\noindent\textbf{Training losses.} As shown in  Figure~\ref{fig_SE}, we adopt a trained structure extraction model from \cite{yi2020unpaired} (referred to as PDG in paper) to produce structure map for the original well-lit portrait, which is treated as pseudo ground-truth for supervising the training of our SE-Net. By respectively denoting our SE-Net and the PDG model as $G_s$ and $G_p$, the training losses of our SE-Net are defined as:
\begin{equation}
\left\{
\begin{aligned}
&\mathcal{L}_{rec} = \Vert{\mathit{G}_s(I_{syn}) - G_p(I)}\Vert_1, \\
&\mathcal{L}_{perceptual} = \mathcal{L}_{LPIPS}(\mathit{G}_s(I_{syn}), G_p(I)), \\
\end{aligned}
\right.
\label{eq3}
\end{equation}
where $\mathcal{L}_{LPIPS}$ refers to the learned perceptual image patch similarity \cite{zhang2018unreasonable}. Besides, we also utilize a discriminator to determine whether the output of SE-Net belongs to the structure image domain, and further define an adversarial loss $\mathcal{L}_{GAN}$ (see supplementary material for details). In summary, the overall loss function for training the SE-Net is given by:
\begin{equation}
\mathcal{L}_{total} = \mathcal{L}_{rec} + \lambda_1\mathcal{L}_{perceptual} + \lambda_2\mathcal{L}_{GAN},
\end{equation}
where we empirically set $\lambda_1 = 0.5$, $\lambda_1 = 0.25$, which produce good results, as demonstrated in Figure \ref{fig:different_structures}.

\subsection{Structure-Guided Portrait Inpainting}
As shown in Figure~\ref{fig_inference}, with structure map, the next step is to inpaint the shadow regions. To this end, we develop a structure-guided portrait inpainting diffusion model.

\vspace{0.5em}
\noindent\textbf{Structure-aware denoising.} As shown in Figure~\ref{fig_train_diff}, we start with a shadow-free portrait as input and then randomly mask some facial regions to create a masked input, followed by a diffusion model to reconstruct the original input by inpainting the masked regions. The inpainting is rooted in the forward diffusion process, where the diffusion model learns a Markovian chain by progressively corrupting the target image $x_0$ (namely the masked input) with Gaussian noise at each timestep $t$, such that it is transformed into a Gaussian distribution (referred to as $\mathcal{N}$ in the following). Specifically, after reparameterization~\cite{ho2020denoising}, our forward process obtains $x_{t}$ from $x_0$ in a single step by:
\begin{equation}
x_t = \sqrt{\bar{\alpha}_t}x_0 + \sqrt{1-\bar{\alpha}_t}\epsilon,\label{x-t}
\end{equation} 
where $\bar{\alpha}_t = \prod_{i=0}^t \alpha_i$ is the cumulative product of forward noise weight parameters $\alpha_i = 1 - \beta_i$, while $\epsilon\sim\mathcal{N}(0,I)$ denotes random Gaussian noise.

As for the reverse denoising process, we use the sampling strategy of DDIM~\cite{song2020denoising} for acceleration. Note, unlike unconditional image generation, our goal is to faithfully reconstruct the original unmasked portrait, the reverse denoising process is thus conditioned on a structure map $S$ and defined as:
\begin{equation}
x_{t-1} = \sqrt{\bar{\alpha}_{t-1}} \left( \frac{x_t - \sqrt{1 - \bar{\alpha}_t} \cdot \mathbf{e}_{t}}{\sqrt{\bar{\alpha}_t}} \right) + \sqrt{1 - \bar{\alpha}_{t-1}} \cdot \mathbf{e}_{t},\label{eq9}
\end{equation} 
where $\mathbf{e}_{t}$ = $\epsilon_\theta(x_t,I_{M},S,M,t)$ refers to the noise estimated by a deep neural network $\epsilon_\theta$, where $I_M$ is the masked input created from a random mask $M$. Following~\cite{ho2020denoising}, the training objective for diffusion model can be simplified as:
\begin{equation}
\mathcal{L}_{diff}(\epsilon) = \Vert \epsilon - \epsilon_\theta(x_t, I_{M}, S, M, t) \Vert^2_2,\label{eq10}
\end{equation}
where $\epsilon$ is the ground truth noise, while $\epsilon_\theta(x_t, I_{M}, S, M, t)$ is the noise predicted by the network. As shown in Figure~\ref{fig_inference}, a facial shadow mask is required at the inference stage, which is a dilated version of the facial shadow detection result from the trained model of \cite{fang2021robust}.

\vspace{0.5em}

\noindent\textbf{Shadow mask refinement.} To enhance the robustness of our method to inaccurate shadow mask, we propose to refine the mask based on the shadow removal result generated by the inpainting diffusion model. Specifically, we first compute the absolute difference between the input image and the shadow removal result, and then perform Otsu-based thresholding to get a refined shadow mask excluding all the non-shadow regions. With the refined mask, we further update the shadow removal result by combining it with the input image, so as to ensure that non-shadow regions in the result are unchanged.   


\begin{figure*}[t]
\centering
\captionsetup[subfigure]{labelformat=empty,skip=1pt}
    \begin{subfigure}[c]{0.12\linewidth}
        \centering
        \includegraphics[width=\textwidth]{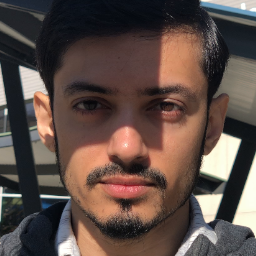} \\ \vspace{1pt}
        \includegraphics[width=\textwidth]{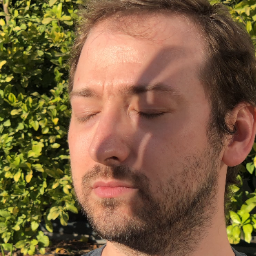}
        \\ \vspace{1pt}
        \includegraphics[width=\textwidth]{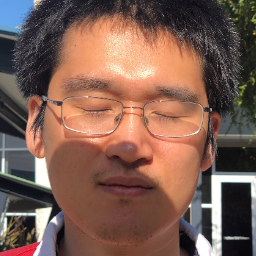}
        \\ \vspace{1pt}
        \includegraphics[width=\textwidth]{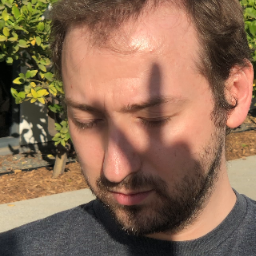}
        \vspace{-5mm}
        \caption{Input}
    \end{subfigure} \hfill
    \begin{subfigure}[c]{0.12\linewidth}
        \centering
        \includegraphics[width=\textwidth]{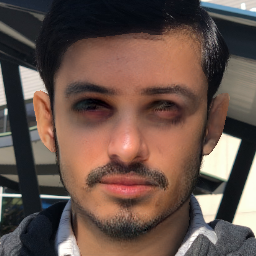} \\ \vspace{1pt}
        \includegraphics[width=\textwidth]{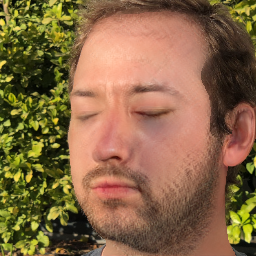}
        \\ \vspace{1pt}
        \includegraphics[width=\textwidth]{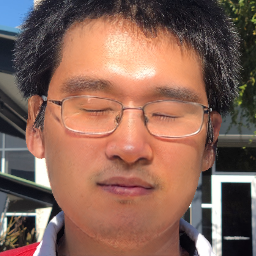}
        \\ \vspace{1pt}
        \includegraphics[width=\textwidth]{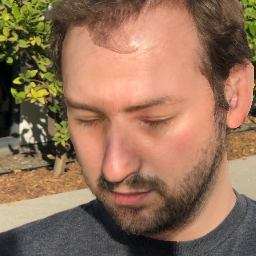}
        \vspace{-5mm}
        \caption{UPSR}
    \end{subfigure} \hfill
    \begin{subfigure}[c]{0.12\linewidth}
        \centering
        \includegraphics[width=\textwidth]{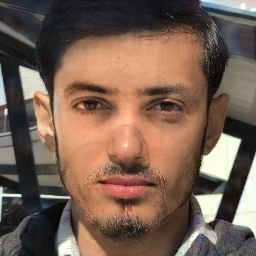} \\ \vspace{1pt}
        \includegraphics[width=\textwidth]{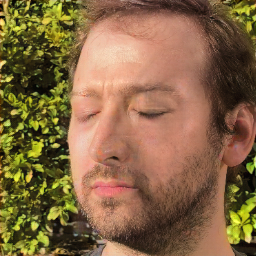}
        \\ \vspace{1pt}
        \includegraphics[width=\textwidth]{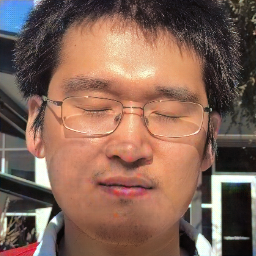}
        \\ \vspace{1pt}
        \includegraphics[width=\textwidth]{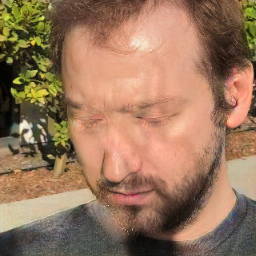}
        \vspace{-5mm}
        \caption{BSR}
    \end{subfigure} \hfill
    \begin{subfigure}[c]{0.12\linewidth}
        \centering
        \includegraphics[width=\textwidth]{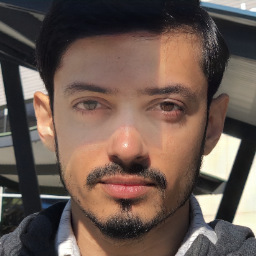} \\ \vspace{1pt}
        \includegraphics[width=\textwidth]{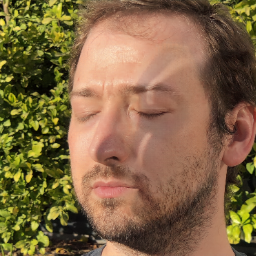}
        \\ \vspace{1pt}
        \includegraphics[width=\textwidth]{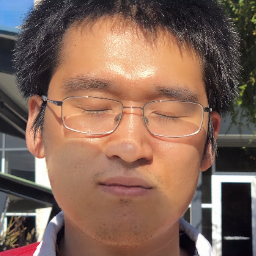}
        \\ \vspace{1pt}
        \includegraphics[width=\textwidth]{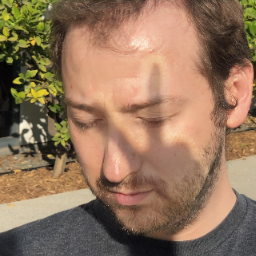}
        \vspace{-5mm}
        \caption{Inpaint4Shadow}
    \end{subfigure} \hfill
    \begin{subfigure}[c]{0.12\linewidth}
        \centering
        \includegraphics[width=\textwidth]{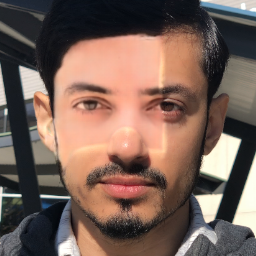} \\ \vspace{1pt}
        \includegraphics[width=\textwidth]{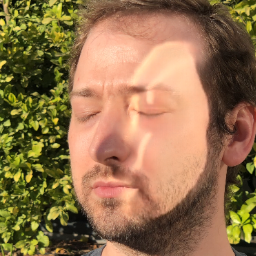}
        \\ \vspace{1pt}
        \includegraphics[width=\textwidth]{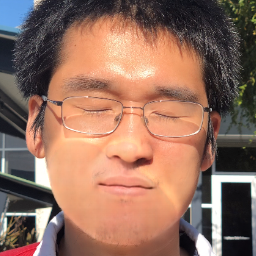}
        \\ \vspace{1pt}
        \includegraphics[width=\textwidth]{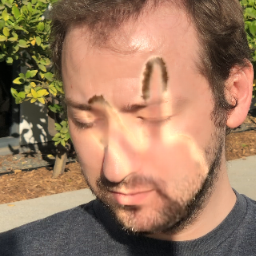}
        \vspace{-5mm}
        \caption{HomoFormer}
    \end{subfigure} \hfill
    \begin{subfigure}[c]{0.12\linewidth}
        \centering
        \includegraphics[width=\textwidth]{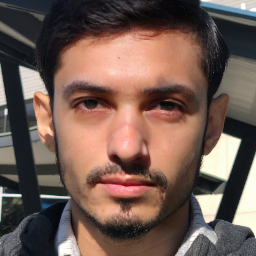} \\ \vspace{1pt}
        \includegraphics[width=\textwidth]{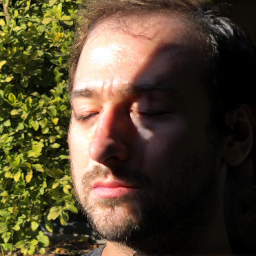}
        \\ \vspace{1pt}
        \includegraphics[width=\textwidth]{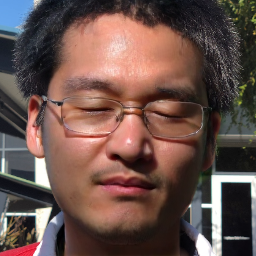}
        \\ \vspace{1pt}
        \includegraphics[width=\textwidth]{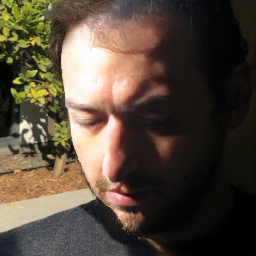}
        \vspace{-5mm}
        \caption{IC-light}
    \end{subfigure} \hfill
    \begin{subfigure}[c]{0.12\linewidth}
        \centering
        \includegraphics[width=\textwidth]{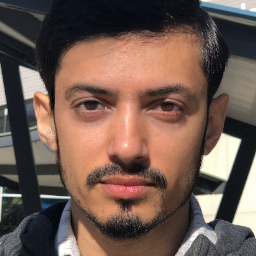} \\ \vspace{1pt}
        \includegraphics[width=\textwidth]{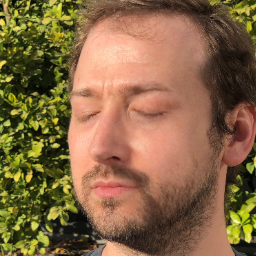}
        \\ \vspace{1pt}
        \includegraphics[width=\textwidth]{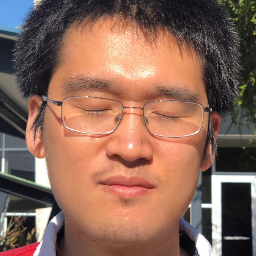}
        \\ \vspace{1pt}
        \includegraphics[width=\textwidth]{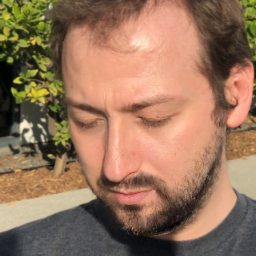}
        \vspace{-5mm}
        \caption{Ours}
    \end{subfigure} \hfill
    \begin{subfigure}[c]{0.12\linewidth}
        \centering
        \includegraphics[width=\textwidth]{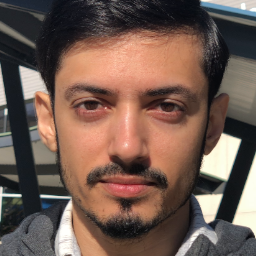} \\ \vspace{1pt}
        \includegraphics[width=\textwidth]{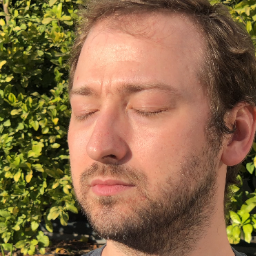}
        \\ \vspace{1pt}
        \includegraphics[width=\textwidth]{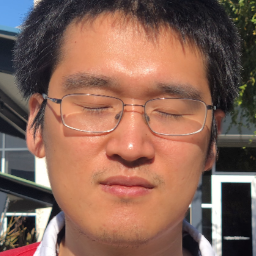}
        \\ \vspace{1pt}
        \includegraphics[width=\textwidth]{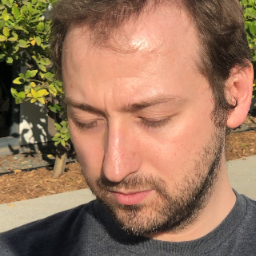}
        \vspace{-5mm}
        \caption{GT}
    \end{subfigure}
    \vspace{-2mm}
     \caption{\textbf{Visual comparison with state-of-the-art methods on test images from the dataset provided in \cite{zhang2020portrait}.}}
\label{fig:com_psm_data}
\end{figure*}

\begin{table*}[t]
  \begin{center}
    \vspace{-2mm}
    \resizebox{\textwidth}{!}{
    \begin{tabular}{lc c cc c cc c c}
        \toprule
        \multirow{2}{*}{Method} & \multicolumn{3}{c}{All} & \multicolumn{3}{c}{Shadow region} & \multicolumn{3}{c}{Non-shadow region}\\
        & SSIM$\uparrow$ & LPIPS$\downarrow$ & RMSE$\downarrow$ & SSIM$\uparrow$ & LPIPS$\downarrow$ & RMSE$\downarrow$ & SSIM$\uparrow$ & LPIPS$\downarrow$ & RMSE$\downarrow$\\
        \midrule
        ShadowDiffusion \cite{guo2023shadowdiffusion} & 0.650 & 0.177 & 38.246 & 0.901 & 0.037 & 15.561 & 0.579 & 0.223 & 52.276\\
        Inpaint4Shadow \cite{li2023leveraging} & 0.766 & 0.094 & 22.709 & 0.910 & 0.033 & 17.250 & 0.708 & 0.164 & 45.456\\
        ShadowFormer \cite{guo2023shadowformer} & 0.772 & 0.086 & 21.413 & 0.926 & 0.027 & 15.288 & 0.723 & 0.149 & 42.341\\
        HomoFormer \cite{xiao2024homoformer} & 0.786 & 0.100 & 21.724 & 0.913 & 0.041 & 16.248 & 0.711 & 0.167 & 42.950\\
        \midrule
        PSM \cite{zhang2020portrait} & 0.782 & 0.074 & - & - & - & - & - & - & -\\
        UPSR \cite{he2021unsupervised} & 0.731 & 0.109 & 26.915 & 0.900 & 0.039 & 20.659 & 0.668 & 0.188 & 45.391\\
        BSR \cite{liu2022blind} & 0.605 & 0.118 & 28.532 & 0.910 & 0.036 & 15.193 & 0.548 & 0.171 & 43.627\\ 
        IC-light~\cite{zhangscaling} & 0.514 & 0.278 & 62.539 & 0.916 & 0.057 & 33.061 & 0.478 & 0.319 & 69.208\\ \midrule
        Ours w/o structure guidance & 0.757 & 0.101 & 23.598 & 0.898 & 0.043 & 18.514 & 0.714 & 0.138 & 39.000\\
        Ours w/ PDG predicted structure & 0.779 & 0.092 & 19.963 & 0.919 & 0.035 & 14.062 & 0.725 & 0.131 & 37.387\\
        Ours w/ data synthesis in PSM  & 0.785 & 0.086 & 19.797 & 0.925 & 0.032 & 13.752 & 0.729 & 0.136 & 38.485\\
        Ours w/o detail-restoration & {0.814} & {0.062} & {18.537} & {0.943} & {0.020} & {12.748} & {0.747} & {0.120} & {36.326}\\
        \midrule
        Ours (full method) & \textbf{0.830} & \textbf{0.056} & \textbf{17.162} &  \textbf{0.973} & \textbf{0.011} & \textbf{10.198} & \textbf{0.778} & \textbf{0.091} & \textbf{34.573}\\
      \bottomrule
    \end{tabular}
    }
    \vspace{-2mm}
    \caption{\textbf{Quantitative comparison on the real-world portrait shadow removal dataset provided in \cite{zhang2020portrait}.} }
    \label{tab1}
  \end{center}
  \vspace{-2mm}
\end{table*}  

\begin{figure*}[t]
\centering
\captionsetup[subfigure]{labelformat=empty,skip=1pt}
    \begin{subfigure}[c]{0.12\linewidth}
        \centering
        \includegraphics[width=\textwidth]{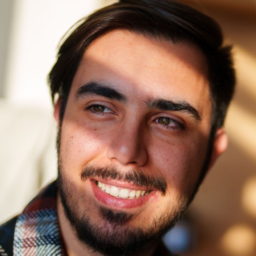} \\ \vspace{1pt}
        \includegraphics[width=\textwidth]{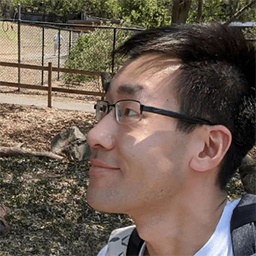} \\ \vspace{1pt}
        \includegraphics[width=\textwidth]{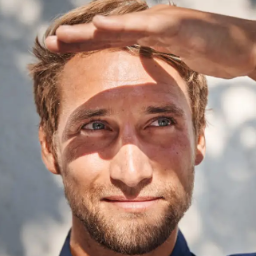} \\ \vspace{1pt}
        \includegraphics[width=\textwidth]{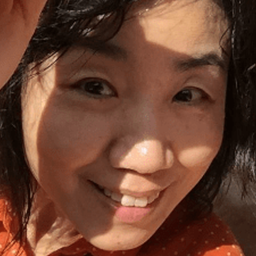} \\
        \vspace{-1mm}
        \caption{Input}
    \end{subfigure} \hfill
    \begin{subfigure}[c]{0.12\linewidth}
        \centering
        \includegraphics[width=\textwidth]{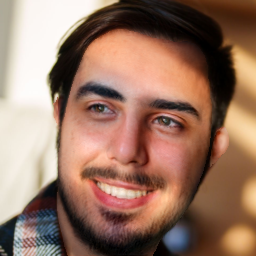} \\ \vspace{1pt}
        \includegraphics[width=\textwidth]{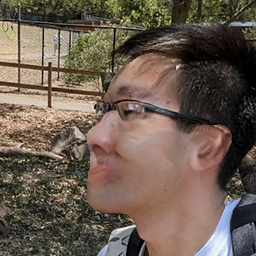} \\ \vspace{1pt}
        \includegraphics[width=\textwidth]{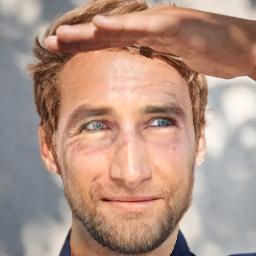}  \\ \vspace{1pt}
        \includegraphics[width=\textwidth]{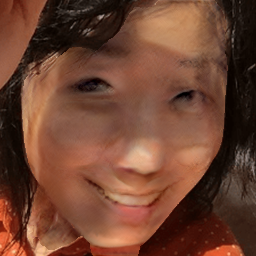}\\
        \vspace{-1mm}
        \caption{UPSR}
    \end{subfigure} \hfill
    \begin{subfigure}[c]{0.12\linewidth}
        \centering
        \includegraphics[width=\textwidth]{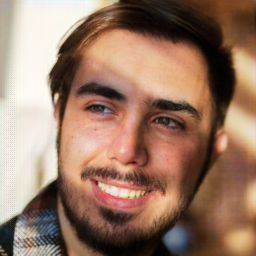} \\ \vspace{1pt}
        \includegraphics[width=\textwidth]{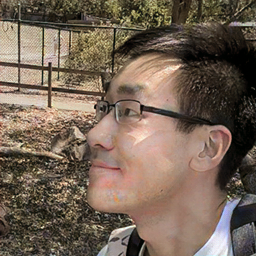} \\ \vspace{1pt}
        \includegraphics[width=\textwidth]{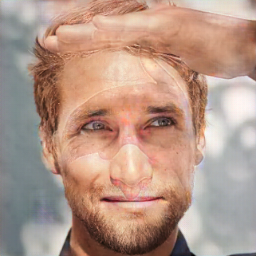}  \\ \vspace{1pt}
        \includegraphics[width=\textwidth]{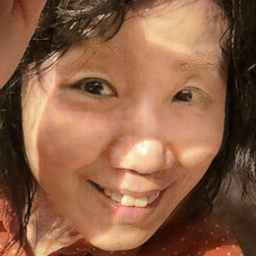}\\
        \vspace{-1mm}
        \caption{BSR}
    \end{subfigure} \hfill
    \begin{subfigure}[c]{0.12\linewidth}
        \centering
        \includegraphics[width=\textwidth]{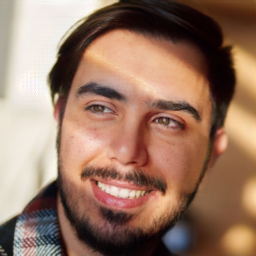} \\ \vspace{1pt}
        \includegraphics[width=\textwidth]{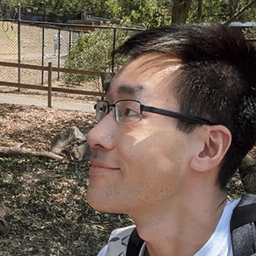} \\ \vspace{1pt}
        \includegraphics[width=\textwidth]{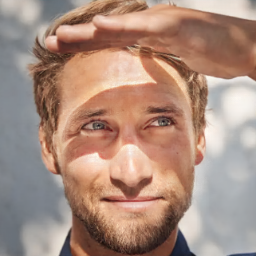} \\ \vspace{1pt}
        \includegraphics[width=\textwidth]{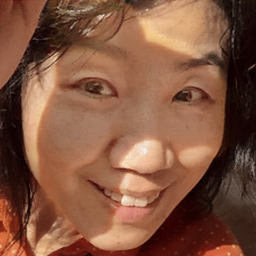} \\
        \vspace{-1mm}
        \caption{Inpaint4Shadow}
    \end{subfigure} \hfill
    \begin{subfigure}[c]{0.12\linewidth}
        \centering
        \includegraphics[width=\textwidth]{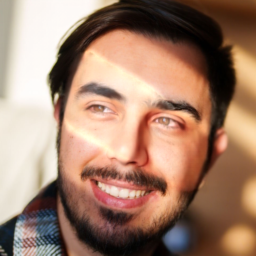} \\ \vspace{1pt}
        \includegraphics[width=\textwidth]{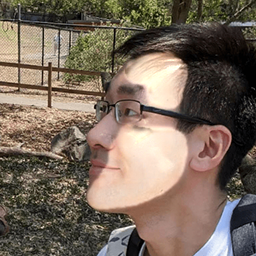} \\ \vspace{1pt}
        \includegraphics[width=\textwidth]{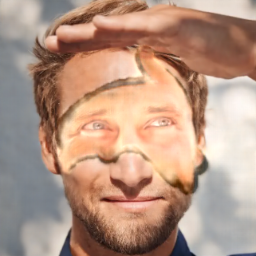} \\ \vspace{1pt}
        \includegraphics[width=\textwidth]{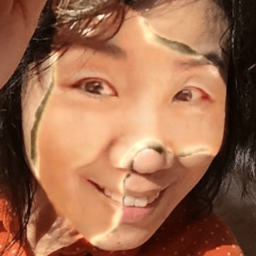}\\
        \vspace{-1mm}
        \caption{HomoFormer}
    \end{subfigure} \hfill
     \begin{subfigure}[c]{0.12\linewidth}
        \centering
        \includegraphics[width=\textwidth]{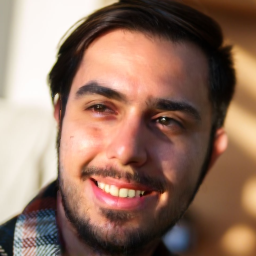} \\ \vspace{1pt}
        \includegraphics[width=\textwidth]{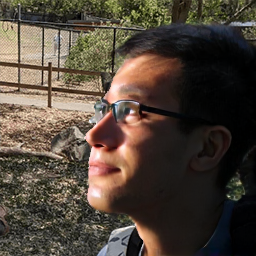} \\ \vspace{1pt}
        \includegraphics[width=\textwidth]{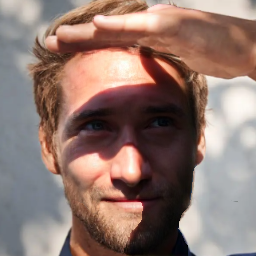}  \\ \vspace{1pt}
        \includegraphics[width=\textwidth]{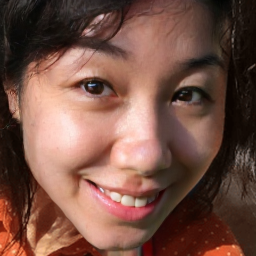} \\
        \vspace{-1mm}
        \caption{IC-light}
    \end{subfigure} \hfill
    \begin{subfigure}[c]{0.12\linewidth}
        \centering
        \includegraphics[width=\textwidth]{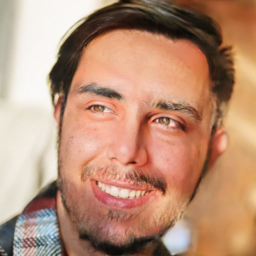} \\ \vspace{1pt}
        \includegraphics[width=\textwidth]{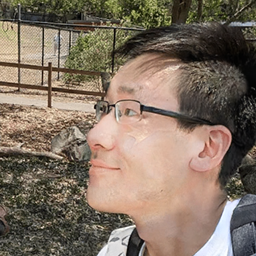} \\ \vspace{1pt}
        \includegraphics[width=\textwidth]{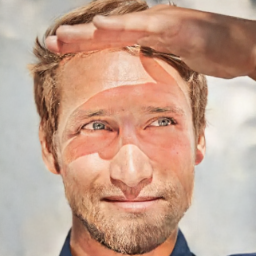}  \\ \vspace{1pt}
        \includegraphics[width=\textwidth]{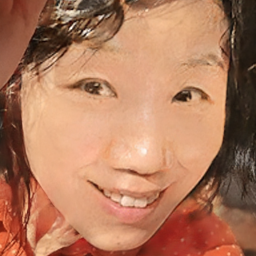}\\
        \vspace{-1mm}
        \caption{ShadowDiffusion}
    \end{subfigure} \hfill
    \begin{subfigure}[c]{0.12\linewidth}
        \centering
        \includegraphics[width=\textwidth]{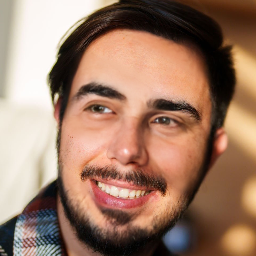} \\ \vspace{1pt}
        \includegraphics[width=\textwidth]{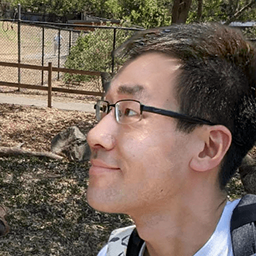} \\ \vspace{1pt}
        \includegraphics[width=\textwidth]{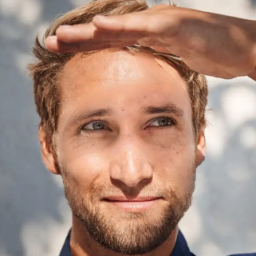} \\ \vspace{1pt}
        \includegraphics[width=\textwidth]{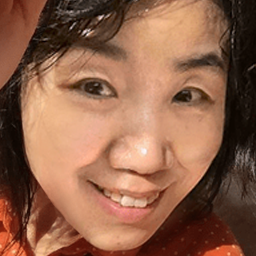} \\
        \vspace{-1mm}
        \caption{Ours}
    \end{subfigure} \hfill
        \vspace{-3mm}
     \caption{\textbf{Visual comparison with state-of-the-art methods on test images from our collected dataset.}}
\label{fig:com_other_data}
\end{figure*}


\subsection{Gradient-Guided Detail Restoration}
As the original PDG model is designed for generating portrait drawing formed by large-scale structures, thus our SE-Net is similarly not so effective in capturing fine-scale portrait details, such as eyelashes, moles, and spots (see the bottom row in Figure~\ref{fig:different_structures} for example). To address this issue, we design a detail restoration diffusion model to refine the shadow removal result generated by the inpainting diffusion model, where we directly take the gradients in the shadow regions as guidance to restore the fine-scale details. Figure~\ref{fig_train_diff} describes how we train the detail restoration diffusion model. Note, gradient-aware denoising process is similar to the structure-aware denoising described above, and the two share the same forward diffusion and reverse denoising formulations described in Equations~\ref{x-t} and~\ref{eq9}, with the latter conditioned on gradient map rather than structure map.

\subsection{Implementation and Training Details}
\textbf{Network architecture.} The SE-Net is an encoder-decoder network, which starts with a flat convolution and two down convolution blocks, followed by nine residual blocks, and finally two up convolution blocks and a final convolution layer. Please see also the supplementary material for visual illustration of the architecture of SE-Net. As for the inpainting diffusion model and the detail-restoration diffusion model, we follow \cite{ho2020denoising} to adopt the same U-Net architecture.

\vspace{0.5em}
\noindent \textbf{Training details.} To train SE-Net, we synthesize 20,000 portrait images with diverse lighting conditions based on the CelebA dataset~\cite{liu2015deep}, and train it for 3 epochs with a batch size of 8 and a learning rate of $1.5 \times 10^{-5}$. As for the inpainting diffusion model, we train it on the FFHQ dataset for 300 epochs with a batch size of 32, and train the detail restoration diffusion model on the same synthetic dataset as the SE-Net for 200 epochs. We build our model on PyTorch and use the Adam optimizer with momentum parameters of (0.9, 0.999) and an initial learning rate of $1 \times 10^{-4}$. Following~\cite{saharia2022image}, we employ the Kaiming initialization~\cite{he2015delving} to initialize the weights of our model and apply 0.9999 Exponential Moving Average (EMA) in all our experiments. We use 2000 diffusion steps and a noise schedule $\beta_t$ linearly increasing from $1 \times 10^{-6}$ to $1 \times 10^{-2}$ for training. During inference, we perform both the inpainting diffusion and detail restoration diffusion for 40 denoising steps, resulting in a total of 80 steps, which takes about 2.5 seconds to process a $512 \times 512$ image on an NVIDIA RTX 4090 GPU.


\section{Experiments}
\textbf{Datasets.} Following previous portrait shadow removal approaches \cite{he2021unsupervised,liu2022blind}, we evaluate our method on the only publicly available real-world paired dataset provided in PSM~\cite{zhang2020portrait}, which contains 9 subjects and 100 shadowed portrait images in varying poses, shadow shapes, illumination conditions, and shadow types. To better examine the effectiveness of our method, we additionally construct a dataset of 200 portraits with diverse facial shadows. Note, we only conduct visual comparison on our collected dataset due to the lack of ground truth shadow removal results.

\vspace{0.5em}
\noindent \textbf{Evaluation metrics.} We use three commonly-used metrics to quantitatively evaluate the performance of our method, including SSIM, LPIPS, and the root mean square error (RMSE) computed in the LAB color space for measuring the similarity between the predicted results and the corresponding ground-truth shadow-free images.

\subsection{Comparison with State-of-the-art Methods}
\textbf{Baselines.} We compare our method with various state-of-the-art shadow removal methods, including four portrait shadow removal and editing methods, i.e., PSM~\cite{zhang2020portrait}, UPSR~\cite{he2021unsupervised}, and BSR~\cite{liu2022blind}, IC-light~\cite{zhangscaling}, as well as four natural image shadow removal methods, i.e., ShadowDiffusion~\cite{guo2023shadowdiffusion}, Inpaint4Shadow~\cite{li2023leveraging}, ShadowFormer~\cite{guo2023shadowformer}, and HomoFormer~\cite{xiao2024homoformer}.  For fair comparison, we produce their results using publicly-available implementation or trained models provided by the authors with recommended parameter setting. Note that, we do not compare with other recent relighting-based shadow editing methods \cite{yoon2024generative,hou2024compose,futschik2023controllable,weir2022deep}, as their codes and trained models are not yet released. 



\vspace{0.5em}
\noindent \textbf{Visual comparison.} We in Figures~\ref{fig:com_psm_data} and \ref{fig:com_other_data} conduct visual comparison between our method and previous shadow removal methods. As shown, shadow removal methods designed for natural images, i.e., Inpaint4Shadow~\cite{li2023leveraging}, HomoFormer~\cite{xiao2024homoformer}, and ShadowDiffusion \cite{guo2023shadowdiffusion} are not well-suited to portraits and tend to induce obvious shadow residuals and appearance distortions. As for the other three portrait shadow removal methods, UPSR~\cite{he2021unsupervised} and BSR~\cite{liu2022blind} fail to preserve the facial details and may also lead to shadow residuals and color distortions, while IC-Light \cite{zhangscaling} struggles to remove the facial shadows due to the inherent difficulty in estimating the ambient light and facial geometry. Please see also the supplementary material for more visual comparison results.


\begin{figure*}[t]
\centering
\captionsetup[subfigure]{labelformat=empty,skip=1pt}
    \begin{subfigure}[c]{0.1385\linewidth}
        \centering
        \includegraphics[width=\textwidth]{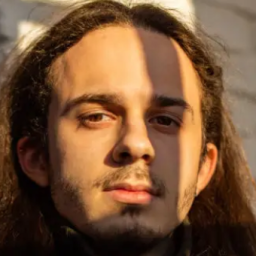} \\ \vspace{1pt}
        \includegraphics[width=\textwidth]{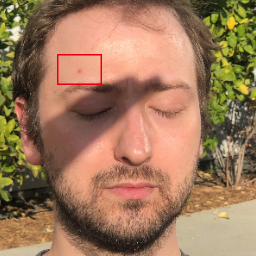} \\
        \vspace{-1mm}
        \caption{}
    \end{subfigure} \hfill
    \begin{subfigure}[c]{0.1385\linewidth}
        \centering
        \includegraphics[width=\textwidth]{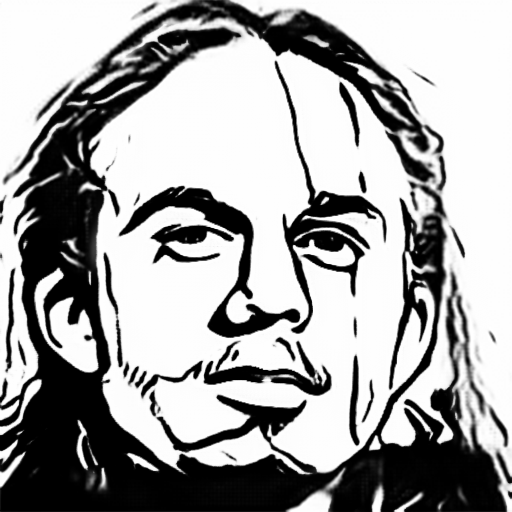} \\ \vspace{1pt}
        \includegraphics[width=\textwidth]{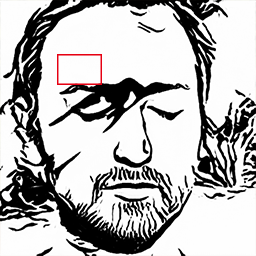}\\
        \vspace{-1mm}
        \caption{}
    \end{subfigure} \hfill
    \begin{subfigure}[c]{0.1385\linewidth}
        \centering
        \includegraphics[width=\textwidth]{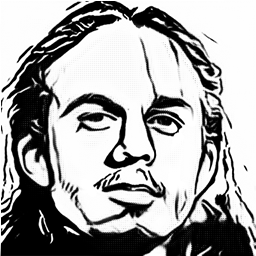} \\ \vspace{1pt}
        \includegraphics[width=\textwidth]{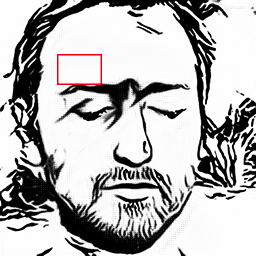}\\
        \vspace{-1mm}
        \caption{}
    \end{subfigure} \hfill
    \begin{subfigure}[c]{0.1385\linewidth}
        \centering
        \includegraphics[width=\textwidth]{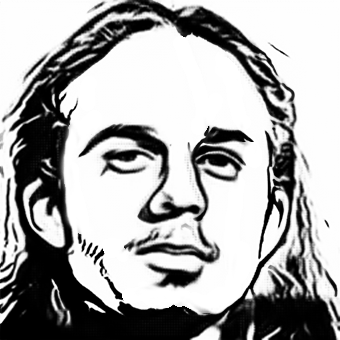} \\ \vspace{1pt}
        \includegraphics[width=\textwidth]{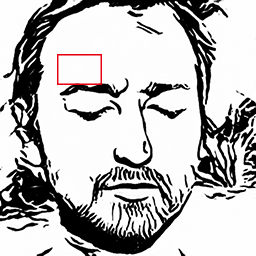}\\
        \vspace{-1mm}
        \caption{}
    \end{subfigure} \hfill
    \begin{subfigure}[c]{0.1385\linewidth}
        \centering
        \includegraphics[width=\textwidth]{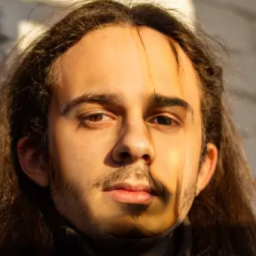} \\ \vspace{1pt}
        \includegraphics[width=\textwidth]{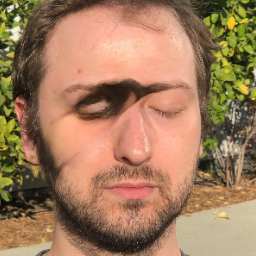}\\
        \vspace{-1mm}
        \caption{}
    \end{subfigure} \hfill
    \begin{subfigure}[c]{0.1385\linewidth}
        \centering
        \includegraphics[width=\textwidth]{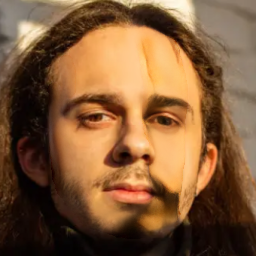} \\ \vspace{1pt}
        \includegraphics[width=\textwidth]{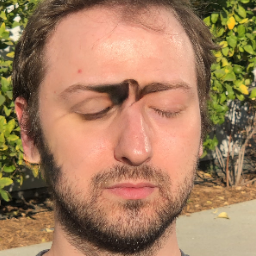}\\
        \vspace{-1mm}
        \caption{}
    \end{subfigure} \hfill
    \begin{subfigure}[c]{0.1385\linewidth}
        \centering
        \includegraphics[width=\textwidth]{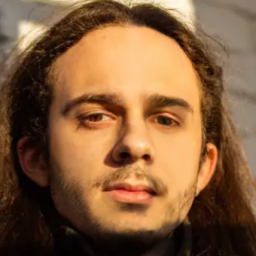} \\ \vspace{1pt}
        \includegraphics[width=\textwidth]{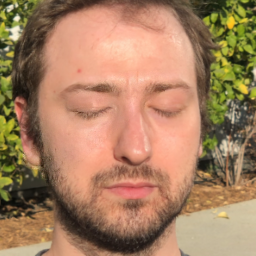}\\
        \vspace{-1mm}
        \caption{}
    \end{subfigure}
    \vspace{-3mm}
\caption{\textbf{Visual comparison of structure maps extracted by different methods and the corresponding shadow removal results.} (a) Input. (b)-(d) are structure maps generated by 
PDG~\cite{yi2020unpaired}, SE-Net trained on data synthesized by the method of PSM~\cite{zhang2020portrait}, and SE-Net trained on our synthesized data. (e)-(g) are shadow removal results produced by our method using the structure maps in (b)-(d), respectively.}
\label{fig:different_structures}
\end{figure*}

\vspace{0.5em}
\noindent \textbf{Quantitative comparison.} In Table~\ref{tab1}, we quantitatively compare our method with other methods in terms of the SSIM, LPIPS, and RMSE metrics. As can be seen, our method outperforms all the compared methods on all the three metrics, demonstrating its effectiveness and superiority in portrait shadow removal. Note, as PSM~\cite{zhang2020portrait} provides only code for synthesizing training data, without releasing the training code and trained model, its quantitative results in Table~\ref{tab1} are as reported in their paper.

\subsection{Ablation Study}
\noindent \textbf{Effectiveness of our shadow-independent structure extraction.} Although the trained model of PDG~\cite{yi2020unpaired} is utilized to train our structure extraction, the structure map produced by PDG~\cite{yi2020unpaired} is not what we expected because it will include the shadow boundaries and lead to unnatural shadow removal results, as shown in Figure~\ref{fig:different_structures} (b)\&(e). By training on our synthesized data with diverse lighting conditions, our structure extraction network (SE-Net) can robustly extract shadow-independent structures, enabling high-quality shadow removal results.

\vspace{0.5em}
\noindent\textbf{Effectiveness of our data synthesis strategy.} To verify the effectiveness of our data synthesis, we alternatively adopt the portrait shadow synthesis strategy introduced in PSM~\cite{zhang2020portrait} to create data on FFHQ for training our SE-Net. As shown in Figure~\ref{fig:different_structures} (c)\&(f) and Table~\ref{tab1}, training on the data synthesized by the strategy in PSM fails to exclude the unwanted shadow boundaries and will degrade the subsequent shadow removal performance, because the strategy only involves simple modifications to brightness, saturation, and subsurface scattering of face, lacking the ability to physically simulate the complex facial lighting conditions.

\vspace{0.5em}
\noindent\textbf{Effect of structure map on shadow inpainting.} We in Figure~\ref{fig:effect_structure} examine how structure map derived from the SE-Net affects the performance of our diffusion-based shadow inpainting. As shown, omitting the structure map, the facial structure details behind shadows are inevitably modified, producing unpredictable portrait inconsistent to the input. In comparison, adopting the structure map as guidance helps generate results with facial structure faithful to the input, validating the necessity of the structure map. 

\vspace{0.5em}
\noindent\textbf{Effect of detail restoration.} We in Figure~\ref{fig:effect_detail} analyze the impact of our detail restoration. As shown, without the restoration step, our shadow removal results exhibit noticeable blurring artifacts around the eyebrows (see 1st row), and fail to faithfully recover the mole in shadow region (see 2nd row). By including the restoration step, we are able to produce high-fidelity shadow removal results with all aforementioned fine-scale details effectively restored, manifesting the effectiveness of our detail restoration.

\begin{figure}[t]
\centering
\captionsetup[subfigure]{labelformat=empty,skip=1pt}
    \begin{subfigure}[c]{0.242\linewidth}
        \centering
        \includegraphics[width=\textwidth]{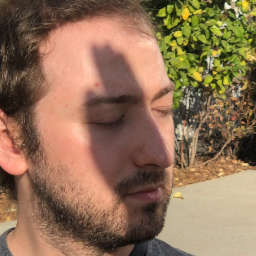} \\ \vspace{1pt}
        \includegraphics[width=\textwidth]{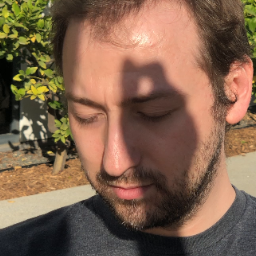} \\ \vspace{-1mm}
        \caption{Input}
    \end{subfigure}
\hfill
    \begin{subfigure}[c]{0.242\linewidth}
        \centering
        \includegraphics[width=\textwidth]{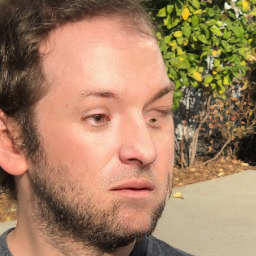} \\ \vspace{1pt}
        \includegraphics[width=\textwidth]{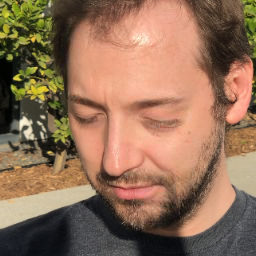} \\ \vspace{-1mm}
        \caption{w/o structure}
    \end{subfigure}
\hfill
    \begin{subfigure}[c]{0.242\linewidth}
        \centering
        \includegraphics[width=\textwidth]{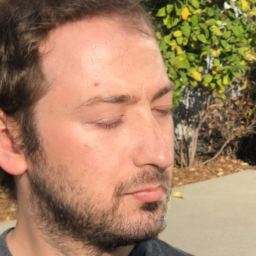} \\ \vspace{1pt}
        \includegraphics[width=\textwidth]{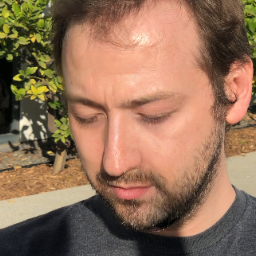} \\ \vspace{-1mm}
        \caption{w/ structure}
    \end{subfigure}
\hfill
    \begin{subfigure}[c]{0.242\linewidth}
        \centering
        \includegraphics[width=\textwidth]{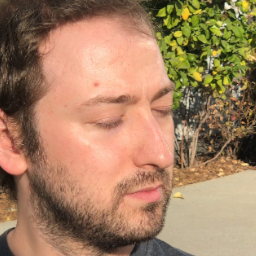} \\ \vspace{1pt}
        \includegraphics[width=\textwidth]{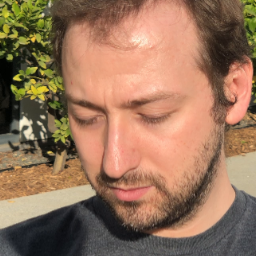} \\ \vspace{-1mm}
        \caption{GT}
    \end{subfigure}
    \vspace{-3mm}
\caption{\textbf{Effect of structure map on shadow inpainting.} }
\label{fig:effect_structure}
\end{figure}

\begin{figure}[t]
\centering
\captionsetup[subfigure]{labelformat=empty,skip=1pt}
    \begin{subfigure}[c]{0.326\linewidth}
        \centering
        \includegraphics[width=\textwidth]{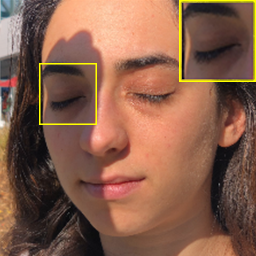} \\ \vspace{1pt}
        \includegraphics[width=\textwidth]{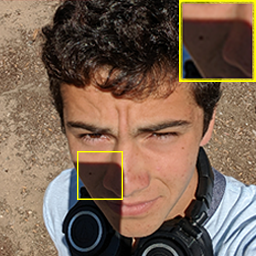} \\ \vspace{-1mm}
        \caption{Input}
    \end{subfigure}
\hfill
    \begin{subfigure}[c]{0.326\linewidth}
        \centering
        \includegraphics[width=\textwidth]{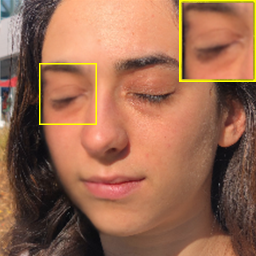} \\ \vspace{1pt}
        \includegraphics[width=\textwidth]{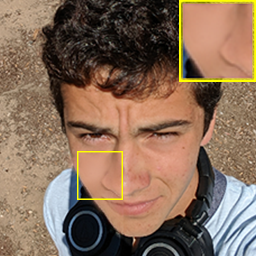} \\
        \vspace{-1mm}
        \caption{w/o detail restoration }
    \end{subfigure}
\hfill
    \begin{subfigure}[c]{0.326\linewidth}
        \centering
        \includegraphics[width=\textwidth]{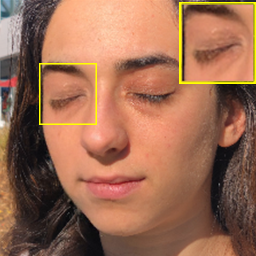} \\ \vspace{1pt}
        \includegraphics[width=\textwidth]{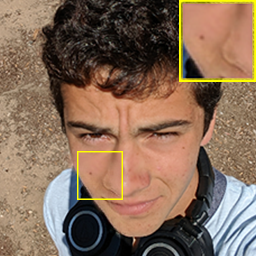} \\
        \vspace{-1mm}
        \caption{w/ detail restoration}
    \end{subfigure}
\hfill
    \vspace{-3mm}
\caption{\textbf{Effect of our detail restoration.} }
\label{fig:effect_detail}
\end{figure}

\subsection{More Analysis}

\vspace{0.5em}
\begin{figure}[t]
\centering
    \captionsetup[subfigure]{labelformat=empty}
    \begin{subfigure}[c]{0.242\linewidth}
        \centering
        \includegraphics[width=\textwidth]{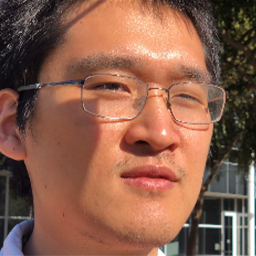} \\ \vspace{-2mm}
        \caption{}
    \end{subfigure}
    \begin{subfigure}[c]{0.242\linewidth}
        \centering
        \includegraphics[width=\textwidth]{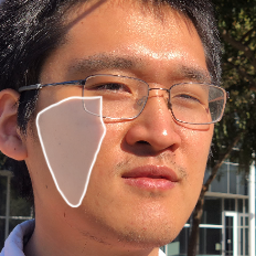} \\\vspace{1pt}
        \includegraphics[width=\textwidth]{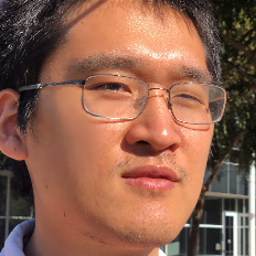} \\
        \caption{}
    \end{subfigure}
    \begin{subfigure}[c]{0.242\linewidth}
        \centering
        \includegraphics[width=\textwidth]{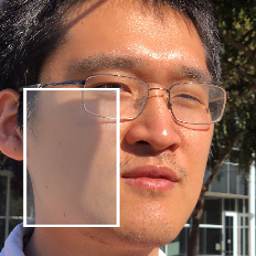} \\\vspace{1pt}
        \includegraphics[width=\textwidth]{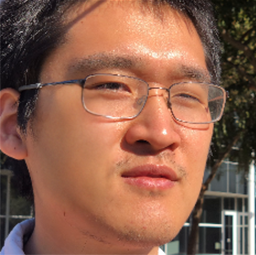} \\
        \caption{}
    \end{subfigure}
    \begin{subfigure}[c]{0.242\linewidth}
        \centering
        \includegraphics[width=\textwidth]{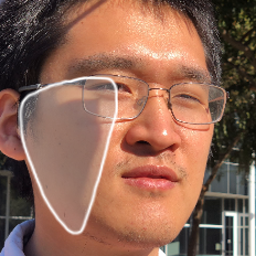} \\\vspace{1pt}
        \includegraphics[width=\textwidth]{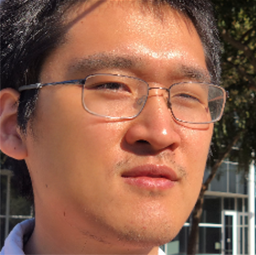} \\
        \caption{}
    \end{subfigure}
    \vspace{-7mm}
\caption{\textbf{Effect of inaccurate shadow masks.} Our method is highly robust to inaccurate masks (see 3rd and 4th columns).}
\label{fig:mask_robustness}
\end{figure}

\begin{figure}[t]
\centering
\captionsetup[subfigure]{labelformat=empty,skip=1pt}
    \begin{subfigure}[c]{0.242\linewidth}
        \centering
        \includegraphics[width=\textwidth]{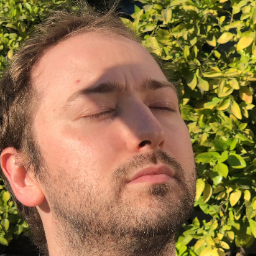} \\ \vspace{1pt}
        \includegraphics[width=\textwidth]{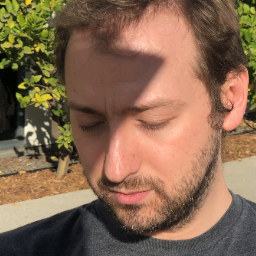} \\ \vspace{-1mm}
        \caption{Input}
    \end{subfigure}
    \hfill
    \begin{subfigure}[c]{0.242\linewidth}
        \centering
        \includegraphics[width=\textwidth]{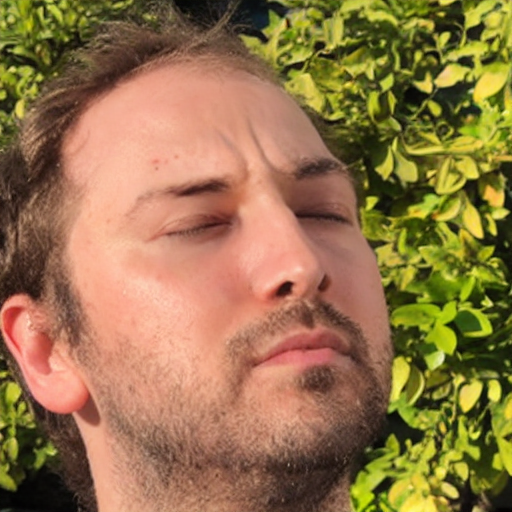} \\ \vspace{1pt}
        \includegraphics[width=\textwidth]{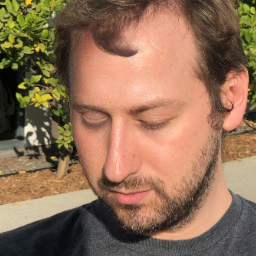} \\ \vspace{-1mm}
        \caption{SD + ControlNet}
    \end{subfigure}
    \hfill
    \begin{subfigure}[c]{0.242\linewidth}
        \centering
        \includegraphics[width=\textwidth]{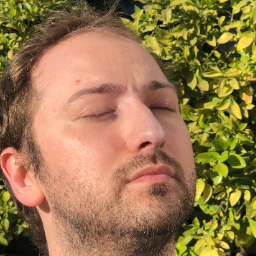} \\ \vspace{1pt}
        \includegraphics[width=\textwidth]{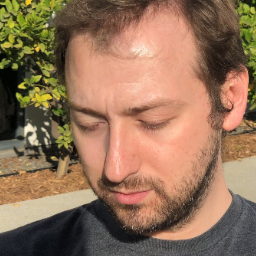} \\ \vspace{-1mm}
        \caption{Ours}
    \end{subfigure}
    \hfill
    \begin{subfigure}[c]{0.242\linewidth}
        \centering
        \includegraphics[width=\textwidth]{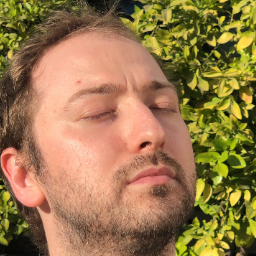} \\ \vspace{1pt}
        \includegraphics[width=\textwidth]{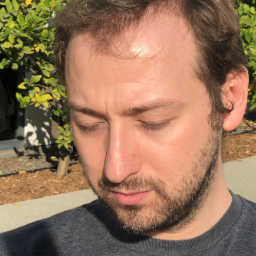} \\ \vspace{-1mm}
        \caption{GT}
    \end{subfigure}
    \hfill
    \vspace{-2mm}
\caption{\textbf{Effect of leveraging different diffusion models for shadow inpainting}. Note, the same structure map as ours is used to produce the shadow inpainting result of ``SD + ControlNet'', and our result here is free of detail restoration.}
\label{fig:other_diff_models}
\end{figure}

\begin{figure}[!t]
\centering
\captionsetup[subfigure]{labelformat=empty,skip=1pt}
    \begin{subfigure}[c]{0.242\linewidth}
        \centering
        \includegraphics[width=\textwidth]{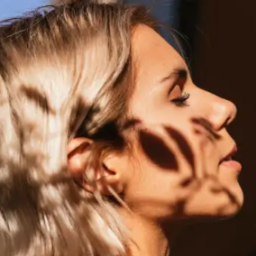} \\ \vspace{-1mm}
        \caption{Input}
    \end{subfigure}
\hfill
    \begin{subfigure}[c]{0.242\linewidth}
        \centering
        \includegraphics[width=\textwidth]{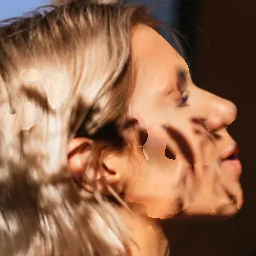} \\
        \vspace{-1mm}
        \caption{UPSR}
    \end{subfigure}
\hfill
    \begin{subfigure}[c]{0.242\linewidth}
        \centering
        \includegraphics[width=\textwidth]{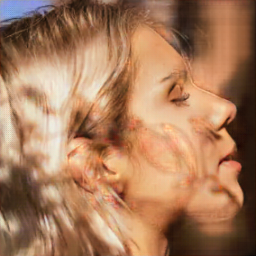} \\
        \vspace{-1mm}
        \caption{BSR}
    \end{subfigure}
\hfill
    \begin{subfigure}[c]{0.242\linewidth}
        \centering
        \includegraphics[width=\textwidth]{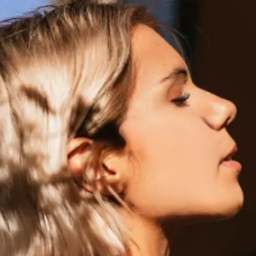} \\
        \vspace{-1mm}
        \caption{Ours}
    \end{subfigure}
    \vspace{-3mm}
\caption{\textbf{Failure case.} Our method may produce unnatural result for complex self-attached shadows (see the eyelid area). }
\label{fig:limit}
\end{figure}
\noindent\textbf{Robustness to inaccurate shadow masks.} Thanks to the design of performing portrait shadow removal via structure-guided inpainting, our method is highly robust to inaccurate input shadow masks, requiring only that the masks cover the shadow regions even if they include additional large-scale non-shadow regions. As demonstrated in Figure~\ref{fig:mask_robustness}, our method produces visually indistinguishable shadow removal results from the three masks, although the last two in the 3rd and 4th columns are inaccurate masks.

\vspace{0.5em}
\noindent\textbf{Different diffusion models for shadow inpainting.} To test how different diffusion models affect the performance of our method, we replace our inpainting diffusion model with a strong pre-trained Stable Diffusion (SD) model (version v1-5 inpainting model) released by \cite{rombach2022high}. To ensure fair comparison, the compared SD model is fine-tuned on our training dataset and injected with the same structure map from SE-Net using ControlNet~\cite{zhang2023adding}. As can be seen in Figure~\ref{fig:other_diff_models}, our trained inpainting diffusion model produces high-fidelity shadow removal results faithful to the inputs, while the SD model incurs noticeable structure tampering. 

\vspace{0.5em}
\noindent\textbf{Effect of our method on natural images.} Our method is also applicable to natural images by replacing the portrait-oriented PDG model~\cite{yi2020unpaired} with a generic edge detection model and also changing the training dataset. We show in the supplementary material that our method can produce competitive shadow removal results for natural images.

\begin{table}[t]
  \begin{center}
 
    \label{tab_time}
    \resizebox{\columnwidth}{!}{
    \begin{tabular}{l c c c c c}
        \toprule
        Method & UPSR & BSR & ShadowDiffusion& IC-light & Ours \\
        \midrule
        Param. & 135M & 72M & 108.79M & - & 185.12M \\
        Time & 192s & 11s & 2.3s & 32s & 9.5s \\
        \bottomrule
    \end{tabular}
    }
    \vspace{-2mm}
    \caption{\textbf{Comparison on the number of parameters (param.) and inference time.} Note, all the inference times here are test for a $512 \times 512$ portrait on an NVIDIA A100 Tensor Core GPU. }
    \vspace{-2mm}
    \label{table:time}
  \end{center}
\end{table}

\vspace{0.5em}
\noindent\textbf{Analysis on computational complexity.} Table~\ref{table:time} compares our inference time with those of other existing portrait specialized or diffusion-based shadow removal methods. As shown, our inference time is more efficient than that of most other methods. Moreover, the number of parameters of our method are comparable to other methods, showing that our superior performance comes mainly from method design.

\vspace{0.5em}
\noindent\textbf{Limitations.} As our method treats shadow removal as diffusion-based inpainting, it gains strong robustness to different types of shadows, especially dark hard shadows that previous methods struggle to handle. However, due to the lack of explicit geometry and physical constraints, it may fail to ensure the overall naturalness when complex self-attached shadows are involved, as shown in Figure~\ref{fig:limit}. 

\vspace{0.5em}
\section{Conclusion}
We have presented a portrait shadow removal approach that allows to robustly produce high-fidelity results. Unlike previous methods, we cast portrait shadow removal as diffusion-based inpainting. To do so, we develop a structure extraction network that is robust to illumination changes to estimate shadow-independent structure map for a given portrait image. The structure map is then fed to an inpainting diffusion model to guide the generation of the shadow-free output. Next, we refine the shadow-free output with a detail restoration diffusion model for impairing the fine-scale facial details that may not captured by the structure map. Extensive experiments show that our method outperforms state-of-the-art methods for portrait shadow removal.

\vspace{0.5em}
\noindent\textbf{Acknowledgement.}This work was supported by the National Natural Science Foundation of China (62471499), the Guangdong Basic and Applied Basic Research Foundation (2023A1515030002).

{
    \small
    \bibliographystyle{ieeenat_fullname}
    \bibliography{main}
}
\end{document}